\documentclass{article}

\usepackage{PRIMEarxiv}

\usepackage[utf8]{inputenc} 
\usepackage[T1]{fontenc}    
\usepackage{hyperref}       
\usepackage{url}            
\usepackage{booktabs}       
\usepackage{amsfonts}       
\usepackage{nicefrac}       
\usepackage{microtype}      
\usepackage{lipsum}
\usepackage{fancyhdr}       
\usepackage{graphicx}       
\graphicspath{{media/}}     

\title{Ablation study of self-supervised learning for image classification}

\author{
  Ilias Papastratis \\
   Aristotle University \\
   Department of Informatics\\
   DMCI\\
  Thessaloniki \\

  \texttt{papastrai@csd.auth.gr} \\

}

\begin{document}
\maketitle

\begin{abstract}
This project focuses on the self-supervised training of convolutional neural networks (CNNs) and transformer networks for the task of image recognition. A simple siamese network with different backbones is used in order to maximize the similarity of two augmented transformed images from the same source image. In this way, the backbone is able to learn visual information without supervision. Finally, the method is evaluated on three image recognition datasets.
\end{abstract}


\section{Introduction}

In this project, a study on  self-supervised training of convolutional neural networks and vision transformers is performed. The self-supervised method \cite{chen2021exploring} with different backbone networks is implemented. The overview of the method is shown in Figure \ref{fig:ssl}. A series of experiments will be conducted to justify the benefits of supervised learning as a prior step to image recognition tasks. Our code is availabe  on GitHub\footnote{\url{https://github.com/iliasprc/SSL-vit-cnn}}.

\subsection{Datasets}

\paragraph{CIFAR-10}  consists of 60000 $32\times 32$ colour images of 10 classes, with 6000 images per class .The 10 different classes represent airplanes, cars, birds, cats, deer, dogs, frogs, horses, ships, and trucks. In total, there are 50000  images and 10000 images on the training and test sets, respectively.
\paragraph{STL-10}
 is an image recognition dataset that is designed  for  unsupervised training of  deep learning  algorithms. It is a similar dataset to CIFAR-10  but each training class has fewer labeled training examples than in CIFAR-10. However, there is  very large set of unlabeled examples that is adopted to train image models in an unsupervised manner and generate self-trained weights. STL-10 has a  higher image resolution of $96\times96$ in contrast to $32\times 32$ of CIFAR-10 images. The dataset contains 100000 unlabeled images, 5000 training images and 8000 images for test.
\paragraph{CelebA} is a face attributes dataset  with more than 200000 face images of large variations on the background and pose. The selected multi-label task for this project is face attributes classification with 40 categories.
\begin{figure}[h]
  \centering
  \includegraphics[width = 0.5\textwidth]{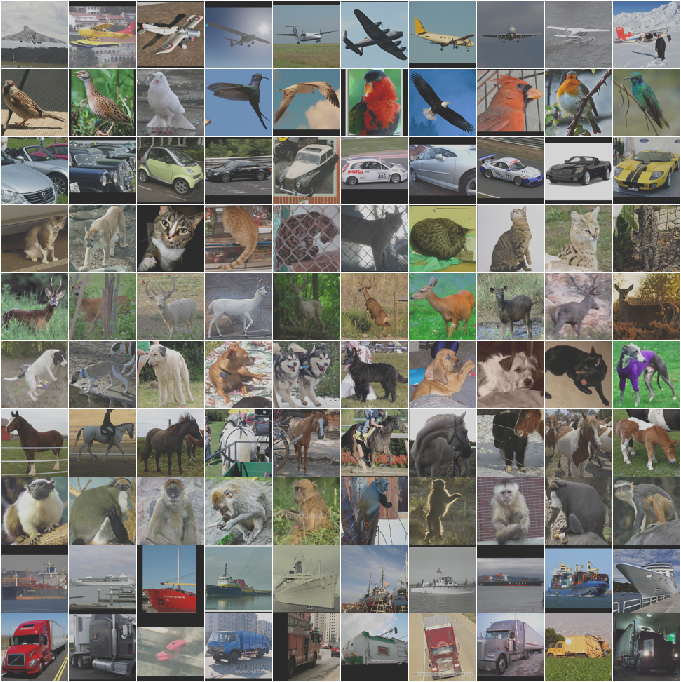}
  \caption{Samples from STL-10 dataset.}
  \label{fig:stl10}
\end{figure}

\subsection{Backbones}
\subsubsection{CNNs}
\paragraph{ResNet-18} is the smaller network of the ResNets \cite{he2016deep} with 18-layers. ResNet-18 contains skip-connections in order to jump over a few layers. This helps the network to avoid the vanishing gradient problem and improve the training speed.

\paragraph{EfficientNet-B0} is the simplest variant of EfficientNets \cite{tan2019efficientnet}. The architecture of EfficientNet-B0 is based on uniformly scaling the dimensions of the network i.e, depth, width and resolution, This is based on the assumption that if the input image is bigger, then the network should be deeper, have an increased receptive field and extract more channels in order to capture better information. 
\subsection{Vision Transformers}

\paragraph{Vision Transformer (ViT)}
  \cite{dosovitskiy2020image} has applied  the Transformer architecture, which was originally implemented for natural language processing, into computer vision tasks and has shown great performance compared to state-of-the-art CNNs.
Each image is split  into fixed-size patches of size $16 \times 16$ which are linearly projected. Then, these projections are linearly embedded and summed with  position  embeddings. The    resulting  sequence  of  vectors  is fed to  a  standard  Transformer encoder. In this project, the ViT-S variant from \cite{rw2019timm} is adopted.

\paragraph{Pooling-based  Vision  Transformer  (PiT)}
\cite{heo2021rethinking} shares a similar architecture with ViT but uses the  dimension reduction principle from CNNs. More specifically, in deeper layers the spatial dimension is reduced with pooling operation while the number of the channels are increased. In this project, the PiT-Ti variant from \cite{rw2019timm} is adopted.
\section{Method}

\subsection{Exploring Simple Siamese Representation Learning}
\begin{figure}[h]
  \centering
  \includegraphics[width = 0.9\textwidth]{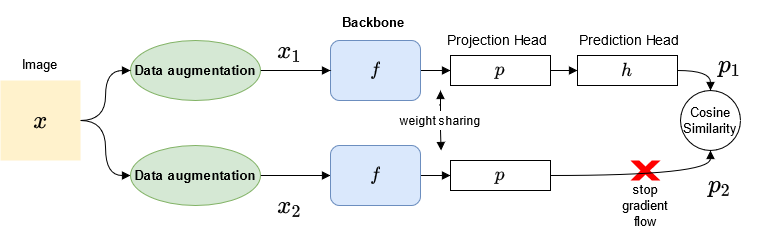}
  \caption{Self-supervised model.}
  \label{fig:ssl}
\end{figure}
The implemented architecture from work \cite{chen2021exploring} (depicted in Figure \ref{fig:ssl}), generates two augmented inputs $x_1,x_2$ from an image $x$. Then, these inputs are fed to the backbone network and the projection head that extracts the visual features. Any CNN or vision Transformer can be adopted as backbone network. The projection head has 3 fully-connected layers with batch normalization applied to each layer except the output layer. Similarly, the prediction head has 2 fully connected layers with batch normalization applied to each layer except the output layer. However, the hidden layer has a reduced dimension by a factor of 4 i.e, has a bottleneck structure. The  output vectors $p_1, p_2, z_1, z_2$ are computed as:
\begin{equation}
        p_1 = h(p(f(x_1))),~~p_2 = h(p(f(x_2))),~~z_1 = p(f(x_1)),~~z_2 = p(f(x_2))
\end{equation}
The negative cosine similarity of two vectors is calculated as:
\begin{equation}
    D(x,y) = -\frac{x}{||x||_2}\frac{y}{||y||_2},
\end{equation}
where $||~||_2$ is the $l_2$ norm.
The network is trained with the symmetric loss function:
\begin{equation}
    L = 0.5*D(p_1, z_2) + 0.5*D(p_2, z_1)
\end{equation}
\section{Implementation details}
\paragraph{Self-supervised learning}
All models are self-supervised trained for 100 epochs with the siamese representation learning model. Adam optimizer is used with an initial learning rate of $1e-3\times \frac{BatchSize}{256}$ with  a  cosine  decay  schedule and a weight decay of $0.00001$. Batch size is set to 64 with gradient accumulation of 8. 
\paragraph{Finetuning} In the classification experiments, Adam optimizer is used with an initial learning rate of $1e-3\times$  and a weight decay of $0.00001$, while batch size is set to 64. Cross-entropy loss is selected for the optimization of the networks except for multi-label classification tasks, where binary cross-entropy loss is used.  
All experiments are conducted on an Nvidia Tesla P100 GPU on the Google Colab Pro platform.

\section{Experimental results}

\begin{table}[t]
\centering
\caption{Experiments  with ResNet-18 on CIFAR-10}
\begin{tabular}{@{}ccccc@{}}

\toprule
 Image size & Pretrain & Dataset  & Validation acc & Test acc       \\ \midrule
 $32\times 32$         & CIFAR-10 & CIFAR-10 & 88.59          & \textbf{88.89} \\
           $32\times 32$         & Random   & CIFAR-10 & 83.87          & 83.12          \\
           $32\times 32$         & No      & CIFAR-10 & 75.68          & 75.23          \\

          \bottomrule
\end{tabular}
\end{table}

\subsection{CIFAR-10 dataset}

At first, a ResNet-18 with different self-supervised settings is evaluated on CIFAR-10 dataset. Due to the image size a variant of ResNet-18 is implemented, which has a different first convolutional layer with smaller kernel size and strides. Initially, the model is trained with random initial weights and achieved an accuracy of $75.23\%$ on the test set. Then, the models are trained self-supervised with random images from a gaussian distribution and finetuned on CIFAR-10. The model achieved $83.12\%$ accuracy justifying that it can learn useful visual information even when it is pretrained with random images. Finally, ResNet-18 is trained unsupervised on the CIFAR-10 dataset in order to learn visual transformation on the same dataset and yielded the best result with an accuracy of $88.89\%$.

\begin{table}[t]
\caption{Self-supervised experiments on STL-10 dataset. Pretrained weights on STL-10 were obtained after self-supervised training for 100 epochs. }
\centering
\begin{tabular}{@{}ccccc@{}}
\toprule
Model            & Pretrained weights &  Test acc  \\ \midrule
ResNet-18             & No       &       65.73           \textbf{} \\
                    & STL-10   &         70.23                   \\
                     & ImageNet &          89.82                 \\
EfficientNet-B0        & No       &       65.30           \textbf{} \\
                        & STL-10   &          69.84                 \\
                      & ImageNet &           95.03                \\ \midrule
                
  ViT           & No       &         53.74         \textbf{} \\
                        & STL-10   &      60.45                     \\
                       & ImageNet &        96.26                   \\
 PiT       & No       &             58.14     \textbf{} \\
                    & STL-10   &   64.21                        \\
                     & ImageNet &                   87.31        \\
                \bottomrule
\end{tabular}
\label{table:stl10}
\end{table}

\subsection{STL-10 dataset}

\begin{figure}[h]
  \centering
  \includegraphics[width = 0.9\textwidth]{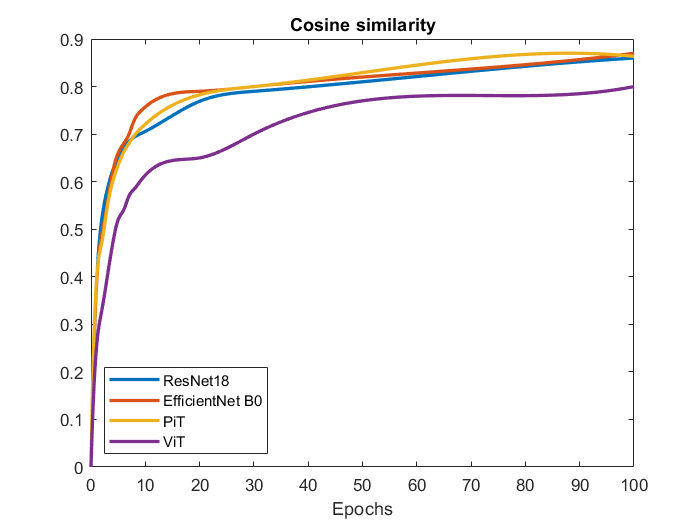}
  \caption{Symmetric cosine similarity loss during self-supervised training.}
  \label{fig:cs}
\end{figure}

The siamese model is now trained with different backbones on the unlabelled set of STL-10. As it can be seen from  Table \ref{table:stl10}, all models benefit from the self-supervised training. ResNet-18 and EfficientNet-B0 achieve about $5\%$ better accuracy (accuracies of $70.23\%$ and $69.84\%$, respectively) on the test set of STL-10 compared to the case, where no pretraining or self-supervised learning is conducted. with random weight initialization. However, when the models are finetuned with pretrained weights on ImageNet, they achieve superior results (accuracies of $89.82\%$ and $95.03\%$, respectively). This might be due to the fact that the self-supervised training is conducted for 100 epochs only. In Figure \ref{fig:cs} the test cosine similarity loss during self-supervised training is depicted. It is shown that ViT converges slower than the rest networks. In general, all backbones must have a cosine similarity close to 1 in order to perform better on classification tasks.
Concerning the vision transformers, they perform relatively worse than CNNs due to the fact that they need larger training datasets. In addition, they converge slower than CNNs during self-supervised training. In general, all models benefit from the self-supervised learning and it is experimentally justified that they can learn useful prior information.
\subsection{CelebA dataset}

The siamese model is  trained with Resnet-18 and PiT backbones on the training set of CelebA dataset. Performance on CelebA is measured with micro and macro average of accuracy and Area under curve (AUC). A macro-average will compute the metric independently for each class and then take the average (hence treating all classes equally), whereas a micro-average will aggregate the contributions of all classes to compute the average metric. As it can be seen, all models benefit slightly from self-supervised training and achieve slightly higher accuracy and AUC performance.


\begin{table}[]
\caption{CelebA  experiments}
\centering
\begin{tabular}{cccccccccc}
\toprule
Model     & Pretrained weights    & \multicolumn{4}{c}{Validation}                        & \multicolumn{4}{c}{Test}                              \\
          &             & \multicolumn{2}{c}{Macro} & \multicolumn{2}{c}{Micro} & \multicolumn{2}{c}{Macro} & \multicolumn{2}{c}{Micro} \\
          &             & Acc         & AUC         & Acc         & AUC         & Acc         & AUC         & Acc         & AUC         \\
          \midrule
ResNet-18 & No Pretrain & 0.5377      & 0.9339      & 0.6522      & 0.9552      & 0.5311      & 0.9301      & 0.6403      & 0.9517      \\

ResNet-18 & CelebA     & 0.5449      & 0.9340      & 0.6549      & 0.9608      & 0.5405      & 0.9310      & 0.6448      & 0.9592      \\
ResNet-18 & Imagenet    & 0.5898      & 0.9391      & 0.6690      & 0.9634      & 0.5826      & 0.9369      & 0.6622      & 0.9617      \\
PiT       &  No Pretrain & 0.5301      & 0.9297      & 0.6412      & 0.9504      & 0.5289      & 0.9265      & 0.6374      & 0.9494     \\
PiT       & CelebA      &   0.5413      & 0.9311     & 0.6579      & 0.9591      & 0.5389      & 0.9303      & 0.6416      & 0.9578         \\
PiT       & Imagenet    &   0.5910          &      0.9440       &      0.6804       &    0.9671         &         0.5841    &     0.9487         &     0.6723        &       0.9623        \\
\bottomrule
\end{tabular}
\end{table}
\section{Conclusion}

In this project, CNNs and vision transformers were evaluated and compared using self-supervised training. Then, they were finetuned and compared on image classification tasks. It was experimentally shown that all models benefit from self-supervised training but they need several iterations of self-supervised training in order to achieve significantly better performance.

\bibliographystyle{unsrt}  
\bibliography{references}

\end{document}